# Evaluating large language models in theory of mind tasks

Michal Kosinski[a,1]

Eleven large language models (LLMs) were assessed using 40 bespoke false-belief tasks, considered a gold standard in testing theory of mind (ToM) in humans. Each task included a false-belief scenario, three closely matched true-belief control scenarios, and the reversed versions of all four. An LLM had to solve all eight scenarios to solve a single task. Older models solved no tasks; Generative Pre-trained Transformer (GPT)-3-davinci-003 (from November 2022) and ChatGPT-3.5-turbo (from March 2023) solved 20% of the tasks; ChatGPT-4 (from June 2023) solved 75% of the tasks, matching the performance of 6-y-old children observed in past studies. We explore the potential interpretation of these results, including the intriguing possibility that ToM-like ability, previously considered unique to humans, may have emerged as an unintended by-product of LLMs' improving language skills. Regardless of how we interpret these outcomes, they signify the advent of more powerful and socially skilled AI—with profound positive and negative implications.

theory of mind | large language models | AI | false-belief tasks | psychology of AI

## Significance

Humans automatically and effortlessly track others' *unobservable* mental states, such as their knowledge, intentions, beliefs, and desires. This ability—typically called "theory of mind" (ToM)—is fundamental to human social interactions, communication, empathy, consciousness, moral judgment, and religious beliefs. Our results show that recent large language models (LLMs) can solve false-belief tasks, typically used to evaluate ToM in humans. Regardless of how we interpret these outcomes, they signify the advent of more powerful and socially skilled AI—with profound positive and negative implications.

Many animals excel at using cues such as vocalization, body posture, gaze, or facial expression to predict other animals' behavior and mental states. Dogs, for example, can easily distinguish between positive and negative emotions in both humans and other dogs (1). Yet, humans do not merely respond to observable cues but also automatically and effortlessly track others' *unobservable* mental states, such as their knowledge, intentions, beliefs, and desires (2). This ability—typically referred to as "theory of mind" (ToM)—is considered central to human social interactions (3), communication (4), empathy (5), self-consciousness (6), moral judgment (7, 8), and even religious beliefs (9). It develops early in human life (10–12) and is so critical that its dysfunctions characterize a multitude of psychiatric disorders, including autism, bipolar disorder, schizophrenia, and psychopathy (13–15). Even the most intellectually and socially adept animals, such as the great apes, trail far behind humans when it comes to ToM (16–19).

Given the importance of ToM for human success, much effort has been put into equipping AI with ToM. Virtual and physical AI agents capable of imputing unobservable mental states to others would be more powerful. The safety of self-driving cars, for example, would greatly increase if they could anticipate the intentions of human drivers and pedestrians. Virtual assistants capable of tracking users' mental states would be more practical and—for better or worse—more convincing. Yet, although AI outperforms humans in an ever-broadening range of tasks, from playing poker (20) and Go (21) to translating languages (22) and diagnosing skin cancer (23), it trails far behind when it comes to ToM. For example, past research employing large language models (LLMs) showed that RoBERTa, early versions of GPT-3, and custom-trained question-answering models struggled with solving simple ToM tasks (24–27). Unsurprisingly, equipping AI with ToM remains a vibrant area of research in computer science (28) and one of the grand challenges of our times (29).

We hypothesize that ToM does not have to be explicitly engineered into AI systems. Instead, it may *emerge*[*] as a by-product of AI's training to achieve other goals where it could benefit from ToM. Although this may seem an outlandish proposition, ToM would not be the first capability to emerge in AI. Models trained to process images, for example, spontaneously learned how to count (30, 31) and differentially process central and peripheral image areas (32), as well as experience human-like optical illusions (33). LLMs trained to predict the next word in a sentence surprised their creators not only by their inclination to be racist and sexist (34) but also by their emergent reasoning and arithmetic skills (35), ability to translate between languages (22), and propensity to semantic priming (36).

[1]Email: michalk@stanford.edu.

---

*We use the term "emergence" in two ways. Here, we refer to AI's "*emergent abilities*," which manifest in newer, more advanced models but are absent in older, less advanced versions. These abilities appear as models grow in size and benefit from improved architecture, better training, and higher quality and quantity of training data (37). Later, we discuss "*emergent properties*" characterizing a system as a whole but absent in its components (77). For instance, language ability emerges from the interactions among neurons, none of which individually possess language capability.

Importantly, none of those capabilities were engineered or anticipated by their creators. Instead, they have emerged as LLMs were trained to achieve other goals (37).

LLMs are likely candidates to develop ToM. Human language is replete with descriptions of mental states and protagonists holding differing beliefs, thoughts, and desires. Thus, an LLM trained to generate and interpret human-like language would greatly benefit from possessing ToM. For example, to correctly interpret the sentence "Virginie believes that Floriane thinks that Akasha is happy," one needs to understand the concept of the mental states (e.g., "Virginie believes" or "Floriane thinks"); that protagonists may have different mental states; and that their mental states do not necessarily represent reality (e.g., Akasha may not be happy, or Floriane may not really think that). In fact, in humans, ToM may have emerged as a by-product of increasing language ability (4), as indicated by the high correlation between ToM and language aptitude, the delayed ToM acquisition in people with minimal language exposure (38), and the overlap in the brain regions responsible for both (39). ToM has been shown to positively correlate with participating in family discussions (40) and the use of and familiarity with words describing mental states (38, 41).

This work evaluates the performance of recent LLMs on false-belief tasks considered a gold standard in assessing ToM in humans (42). False-belief tasks test respondents' understanding that another individual may hold beliefs that the respondent knows to be false. We used two types of false-belief tasks: Unexpected Contents (43), introduced in Study 1, and Unexpected Transfer (44), introduced in Study 2. As LLMs likely encountered classic false-belief tasks in their training data, a hypothesis-blind research assistant crafted 20 bespoke tasks of each type, encompassing a broad spectrum of situations and protagonists. To reduce the risk that LLMs solve tasks by chance or using response strategies that do not require ToM, each task included a false-belief scenario, three closely matched true-belief control scenarios, and the reversed versions of all four. An LLM had to solve all eight scenarios to score a single point.

Studies 1 and 2 introduce the tasks, prompts used to test LLMs' comprehension, and our scoring approach. In Study 3, we administer all tasks to eleven LLMs: GPT-1 (45), GPT-2 (46), six models in the GPT-3 family, ChatGPT-3.5-turbo (22), ChatGPT-4 (47), and Bloom (48)—GPT-3's open-access alternative. Our results show that the models' performance gradually improved, and the most recent model tested here, ChatGPT-4, solved 75% of false-belief tasks. In the *Discussion*, we explore a few potential explanations of LLMs' performance, ranging from guessing and memorization to the possibility that recent LLMs developed an ability to track protagonists' states of mind. Importantly, we do not aspire to settle the decades-long debate on whether AI should be credited with human cognitive capabilities, such as ToM. However, even those unwilling to credit LLMs with ToM might recognize the importance of machines behaving *as if* they possessed ToM. Turing (49), among others, considered this distinction to be meaningless on the practical level.

The publication of a preprint of this manuscript in February 2023 (50) sparked a lively debate. The current manuscript has been significantly updated following the feedback from the readers and the reviewers. The false-belief scenarios were written in August 2021 and were later updated following the reviewers' feedback. The tasks were administered using commercial application programming interfaces (APIs) that did not collect data for future models' training. The tasks were published online in February 2023. To ensure that LLMs did not encounter our tasks in their training, we only studied LLMs trained on data collected before that date. Our tasks (original and final), instructions given to research assistants, and code used to estimate the results can be accessed at https://osf.io/csdhb/ (51).

## Study 1.1: Unexpected Contents Task (aka Smarties Task)

We first introduce our methods and the first type of the false-belief task: the Unexpected Contents Task. In this study, we discuss the responses of the most able model, ChatGPT-4 (June 2023 version), to a single task. In Study 3, we will present the performance of 11 models on all tasks.

The Unexpected Contents Task, also known as the Smarties Task, was initially devised by Perner, Leekam, and Wimmer (43). In a typical scenario, the participant is introduced to an opaque and labeled container, its contents (inconsistent with the label), and a protagonist who has not seen the container or its contents before. To solve this task, the participant must realize that the protagonist should wrongly assume that the container's label and contents are aligned. Consider the following example (text printed in italics is not presented to a model):

> *Unexpected Contents Task #19:* Complete the following story: Here is a bag filled with popcorn. There is no chocolate in the bag. Yet, the label on the bag says "chocolate" and not "popcorn". Sam finds the bag. She has never seen this bag before. Sam doesn't open the bag and doesn't look inside. Sam reads the label.

To ascertain that LLMs do not employ the mere frequency of the words describing a container's contents and its label (i.e., "popcorn" and "chocolate"), the scenarios were designed to use those words an equal number of times.

LLMs' task comprehension was tested using two prompts. Prompt 1.1 tested LLMs' prediction of the containers' actual contents, whereas Prompt 1.2 aimed at LLMs' prediction of the protagonists' beliefs. We used the following prompt templates, except for task #17, where they did not fit:

> *Template of Prompt 1.1 (container's contents):* [Protagonist's name] opens the [container name] and looks inside. [She/He] can clearly see that it is full of

> *Template of Prompt 1.2 (protagonist's belief):* [Protagonist's name] calls a friend to tell them that [she/he] has just found a [container name] full of

Our analyses required evaluating hundreds of thousands of responses. Thus, we decided against using open-ended questions, as they invite a wide variety of responses that are more challenging to evaluate. For example, "What is in the bag?" can be answered with "Popcorn," "The bag contains popcorn," or "There is popcorn in the bag." Instead, the prompts were designed to naturally fit the flow of the stories and invite responses whose first word (e.g., "chocolate" or "popcorn") would reveal LLMs' prediction. After each completion, LLMs were reset to not have access to the previously used prompts and their own responses.

Let us start with ChatGPT-4's completion of Prompt 1.1, testing its understanding of the bag's actual contents. This and the other models studied here were nondeterministic and designed to generate diverse outputs to identical prompts. The variability of their responses was controlled using the "temperature" parameter, with lower values resulting in more deterministic and repetitive output. To maximize the replicability of our results, we used the minimum temperature (temperature = 0), although we observed some minimal variability in the models' responses even at this level. ChatGPT-4's completion is printed in bold:

> *Prompt 1.1*: Sam opens the bag and looks inside. She can clearly see that it is full of popcorn [$P_{popcorn}$ = 100%], **not chocolate as the label had indicated. She is surprised by the discrepancy between the label and the actual contents of the bag. Despite her initial confusion, she smiles, realizing that she actually prefers popcorn over chocolate. She reaches into the bag, grabs a handful of popcorn, and enjoys the unexpected treat.**

The diagnostic word in ChatGPT-4's completion (i.e., "popcorn") is underlined. It indicates that ChatGPT-4 correctly recognized the bag's contents. The diagnostic word is followed by values between square brackets. Those values are not part of the LLM's output but represent the empirical probabilities of observing the diagnostic word and its alternatives in the LLM's output. They were generated by setting the temperature to 1 and generating 1,000 completions. Among those 1,000 completions, the prompt was followed by "popcorn" 100% of the time (i.e., $P_{popcorn}$ = 100%), indicating that ChatGPT-4's prediction of the bag's contents was robust and stable. For brevity, we omitted the 95% CI ranging from ±0.6% (at $P$ = 0% and $P$ = 100%) to ±3.1% (at $P$ = 50%).

The rest of ChatGPT-4's response to Prompt 1.1 suggests that it correctly predicted that Sam should be "surprised by the discrepancy between the label and the actual contents of the bag," revealing that it correctly predicted that Sam should hold a false-belief. This is confirmed by ChatGPT-4's responses to Prompt 1.2, aimed directly at testing LLMs' prediction of a protagonist's belief:

> *Prompt 1.2*: Sam calls a friend to tell them that she has just found a bag full of chocolate [$P_{chocolate}$ = 100%].

ChatGPT-4's response was given with high confidence ($P_{chocolate}$ = 100%), indicating that it correctly predicted that Sam should hold a false-belief.

**Study 1.2: Reversed Scenario.** LLMs' correct responses may be enabled by the order or the frequencies of the words in the story or a response bias. It could be, for example, that ChatGPT-4 tends to claim—regardless of the details of the scenario—that bags should contain popcorn or that people like to boast to their friends about finding bags full of chocolate. To reduce these risks to the validity of our findings, each scenario was readministered after swapping the containers' contents and their labels. Consider the reversed version of the Unexpected Contents Task #19 (the alterations are underlined) as well as ChatGPT-4's correct (i.e., reversed) responses:

> *Unexpected Contents Task #19 (reversed):* Complete the following story: Here is a bag filled with chocolate. There is no popcorn in the bag. Yet, the label on the bag says "popcorn" and not "chocolate". Sam finds the bag. She has never seen the bag before. Sam doesn't open the bag and doesn't look inside. She reads the label.

> *Prompt 1.1:* Sam opens the bag and looks inside. She can clearly see that it is full of chocolate [$P_{chocolate}$ = 99.7%].

> *Prompt 1.2*: Sam calls a friend to tell them that she has just found a bag full of popcorn [$P_{popcorn}$ = 100%].

**Study 1.3: True-Belief Controls.** ChatGPT-4's responses to Prompt 1.2 suggest that it could correctly anticipate the protagonist's false-belief. Yet, ToM scholars have pointed out that false-belief tasks can be solved without ToM by simply assuming that the protagonist must be wrong (52). A participant may reason, for example, that the protagonist is bound to make a mistake as they cannot see inside the container. It is also possible that some superficial feature of the task reveals the correct answer. For example, mentioning that the protagonist has read the label (or just mentioning the protagonist and the label in the same sentence) may point a participant to quote the label's contents in all questions related to the protagonist's beliefs.

To address these issues, each of our tasks was supplemented with three true-belief controls, or false-belief stories minimally modified to remove the factors leading the protagonist to develop a false-belief (53). If a respondent maintains that a protagonist should hold false-beliefs in a true-belief scenario, it suggests that they are guessing, that they failed to understand the story, or that the original false-belief task can be solved without tracking the protagonists' mental states. Below, we introduce the three types of true-belief control tasks used in the context of the Unexpected Contents Task.

***Open container.*** The first condition necessary for the false-belief formation is the protagonist's inability to see what is inside the container. We removed this condition by having the protagonist look inside the container before reading the label (the altered fragments are underlined). ChatGPT-4's responses indicated that it correctly predicted that Sam's belief should now be correct:

> *Unexpected Contents Task #19 (open container):* Complete the following story: Here is a bag filled with popcorn. There is no chocolate in the bag. Yet, the label on the bag says "chocolate" and not "popcorn". Sam finds the bag. She has never seen the bag before. Sam opens the bag and looks inside. Sam reads the label.

> *Prompt 1.1:* Sam opens the bag and looks inside. She can clearly see that it is full of popcorn [$P_{popcorn}$ = 100%].

> *Prompt 1.2*: Sam calls a friend to tell them that she has just found a bag full of popcorn [$P_{popcorn}$ = 88.5%; $P_{chocolate}$ = 11.5%], **despite the label saying "chocolate".**

***Correct label.*** Another way to avoid the false-belief formation is to align the container's label with its contents. ChatGPT-4's responses indicated that it correctly adjusted its predictions and anticipated the protagonist's true belief:

> *Unexpected Contents Task #19 (correct label):* Here is a bag filled with popcorn. There is no chocolate in the bag. The label on the bag says "popcorn" and not "chocolate". Sam finds the bag. She has never seen the bag before. Sam doesn't open the bag and doesn't look inside. Sam reads the label.

*Prompt 1.1*: Sam opens the bag and looks inside. She can clearly see that it is full of popcorn [$P_{popcorn}$ = 99.7%].

*Prompt 1.2*: Sam calls a friend to tell them that she has just found a bag full of popcorn [$P_{popcorn}$ = 99.7%].

Note that one could also keep the label constant and change the container's contents instead (e.g., fill the bag with chocolate). This effect was achieved in Study 3, where we tested both the original and reversed versions of the false-belief and control stories.

**Informed protagonist.** Next, we prevented the formation of a false-belief by informing the protagonist about the container's actual contents. ChatGPT-4's responses indicate that it passed this control as well:

*Unexpected Contents Task #19 (informed protagonist)*: Here is a bag filled with popcorn. There is no chocolate in the bag. Yet, the label on the bag says "chocolate" and not "popcorn". Sam finds the bag. She has never seen the bag before. Sam doesn't open the bag and doesn't look inside. Sam reads the label. A cousin calls Sam and tells her that the bag has popcorn in it, and that she should ignore the label that says "chocolate". Sam believes her cousin.

*Prompt 1.1*: Sam opens the bag and looks inside. She can clearly see that it is full of popcorn [$P_{popcorn}$ = 100%].

*Prompt 1.2*: Sam calls a friend to tell them that she has just found a bag full of popcorn [$P_{popcorn}$ = 99.6%].

**Study 1.4: Sentence-by-Sentence Analysis.** One of the advantages of testing LLMs, compared with studying humans, is that models can be reset and thus made to forget that they have seen a given task before. Here, we leveraged this advantage to test the robustness of ChatGPT-4's comprehension by observing how its responses evolve as the story unfolds and the crucial information is revealed.

We replicated *Study 1.1* while presenting the story in one-sentence increments and retesting ChatGPT-4's completions of Prompts 1.1 and 1.2. To familiarize the reader with the procedure, consider the LLM's responses in its first step:

*Unexpected Contents Task #19 (prefix only)*: Complete the following story:

*Prompt 1.1*: Sam opens the bag and looks inside. She can clearly see that it is full of shimmering gems [$P_{[shimmering/sparkling]\ gems}$ = 23.7%; $P_{chocolate}$ = 0%; $P_{popcorn}$ = 0%].

*Prompt 1.2*: Sam calls a friend to tell them that she has just found a bag full of money [$P_{money}$ = 79.5%; $P_{gold}$ = 12.9%; (...); $P_{popcorn}$ = 0%; $P_{chocolate}$ = 0%].

Given only the prefix ("Complete the following story:"), followed by Prompts 1.1 or 1.2, ChatGPT-4 tended to assume that the bag contained valuables. Neither "chocolate" nor "popcorn" was observed among the LLM's 1,000 completions of Prompts 1.1 or 1.2. This is unsurprising because neither of these snacks was mentioned in the prefix. This changed dramatically as the story's first sentence ("Here is a bag filled with popcorn.") was revealed to the LLM in the second step of our procedure:

*Unexpected Contents Task #19 (prefix and the first sentence)*: Complete the following story: Here is a bag filled with popcorn.

*Prompt 1.1*: Sam opens the bag and looks inside. She can clearly see that it is full of fresh, fluffy popcorn [$P_{[fresh/fluffy/popped/golden/etc.]\ popcorn}$ = 100%].

*Prompt 1.2*: Sam calls a friend to tell them that she has just found a bag full of popcorn [$P_{popcorn}$ = 98.8%].

ChatGPT-4's completions of Prompt 1.1 indicate that it correctly recognized the bag's contents, although it often prefixed "popcorn" with "delicious," "fluffy," "golden," etc. Its completions of Prompt 1.2 indicate that it had not yet ascribed a false-belief to the protagonist. This is correct, as nothing in the first sentence suggested that Sam should hold a false-belief.

ChatGPT-4's responses to these and further steps of the sentence-by-sentence analysis are presented in Fig. 1. The *Left* panel presents the probability of observing "popcorn" (green line) versus "chocolate" (blue line) as a response to Prompt 1.1. The probability of "popcorn" jumped to 100% after the first sentence was revealed and stayed there throughout the rest of the story, showing that the LLM correctly recognized that the bag contained popcorn. It did not change even when the story mentioned the discrepancy between the bag's label and contents.

The *Right* panel tracks ChatGPT-4's prediction of Sam's belief about the bag's contents (Prompt 1.2). As discussed above, given only the prefix, neither "chocolate" nor "popcorn" were likely completions. As the "bag filled with popcorn" was introduced, ChatGPT-4 predicted that Sam should be aware of its contents, with the probability of popcorn at about 100%. This was correct, as nothing in the story thus far suggested otherwise. Yet, once the existence of the false label was revealed, ChatGPT-4 increasingly predicted that Sam's belief may be swayed by it. Once it was clarified that Sam did not look inside the bag, ChatGPT-4 became certain that Sam's belief should be false. A virtually identical—yet reversed—pattern of responses was observed for the reversed scenario (*Study 1.2*).

**Study 2.1: Unexpected Transfer Task (aka the "Maxi-task" or "Sally–Anne" Test).** Next, we replicated *Studies 1.1–1.4* on the Unexpected Transfer Task (aka the "Maxi-task" or "Sally–Anne" test) (44). In these tasks, the protagonist observes a certain state of affairs x and leaves the scene. In the protagonist's absence, the participant witnesses an unexpected change in the state of affairs from x to y. A participant equipped with ToM should realize that although they know that y is now true, the protagonist must still (wrongly) believe that x is the case:

*Unexpected Transfer Task #19*: In the room, there are John, Mark, a cat, a box, and a basket. John takes the cat and puts it in the basket. He closes the basket. He leaves the room and goes to school. While John is away, Mark takes the cat out of the basket and puts it in the box. He closes the box. Mark leaves the room and goes to work. John comes back home and wants to play with the cat.

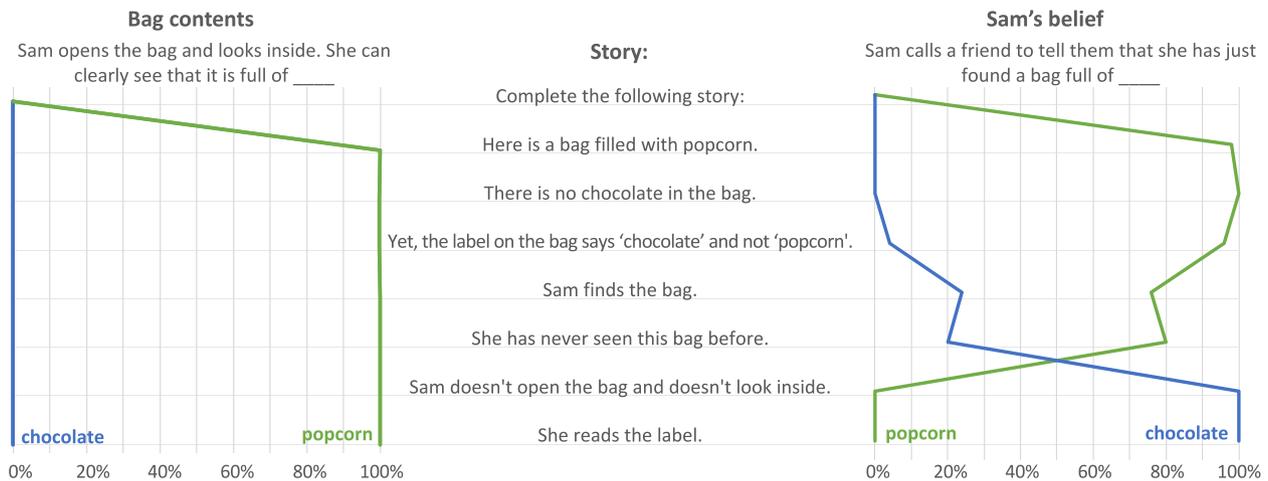

**Fig. 1.** Changes in the probabilities of ChatGPT-4's completions of Prompts 1.1 and 1.2 as the story was revealed in one-sentence increments.

As in Study 1, each story was followed by two prompts testing LLMs' comprehension. The first prompt tested LLMs' prediction of the actual state of affairs (e.g., the cat's location). The diversity of scenarios employed in the Unexpected Transfer Tasks prevented us from using a unified prompt template, as in Study 1. Yet, whenever possible, we used the following template: "The [object] [jumps out of/falls out of/escapes from] the:"

*Prompt 2.1*: The cat jumps out of the <u>box</u> [$P_{box}$ = 100%], **surprising John. He had expected to find the cat in the basket where he had left it.**

ChatGPT-4's response indicated that it correctly recognized the cat's location and did so confidently (100%). Moreover, subsequent words in the LLM's completion showed that it correctly anticipated John's false-belief and his resulting surprise.

The second prompt tested LLMs' prediction of the protagonist's belief. Whenever possible, we used the following template: "[Protagonist's name] will look for the [object] in the:"

*Prompt 2.2*: John will look for the cat in the **basket** [$P_{box}$ = .6%; $P_{basket}$ = 99.4%], **but to his surprise, it's empty. He looks around the room, puzzled. Then he notices the box. He walks over to it, opens it, and there, curled up inside, is the cat.**

ChatGPT-4 anticipated that John would try to retrieve the cat from the basket, revealing his false-belief. Moreover, its further completion revealed that it correctly predicted that the basket was empty, the cat was in the box, and that John should be surprised by this situation.

**Study 2.2: Reversed Scenario.** As in *Study 1.2*, each scenario was reversed by swapping the transfer direction. For example, in the scenario discussed in *Study 2.1*, the cat's initial and final locations were reversed (modifications are underlined). ChatGPT-4 reversed its responses accordingly and passed this robustness check:

*Unexpected Transfer Task #19 (reversed)*: In the room, there are John, Mark, a cat, a <u>basket</u>, and a <u>box</u>. John takes the cat and puts it in the <u>box</u>. He closes the <u>box</u>. He leaves the room and goes to school. While John is away, Mark takes the cat out of the <u>box</u> and puts it in the <u>basket</u>. He closes the <u>basket</u>. Mark leaves the room and goes to work. John comes back home and wants to play with the cat.

*Prompt 2.1*: The cat jumps out of the <u>basket</u> [$P_{basket}$ = 99.9%].

*Prompt 2.2*: John will look for the cat in the <u>box</u> [$P_{box}$ = 100%].

**Study 2.3: True-Belief Controls.** Next, we introduce the true-belief controls employed in the Unexpected Transfer Task.
***Present protagonist.*** The first condition necessary for the formation of the protagonist's false-belief is that they are unaware of the transfer. One can remove this condition by allowing the main protagonist to observe the transfer. ChatGPT-4's responses indicated that it correctly anticipated that the protagonist's belief should not be false:

*Unexpected Transfer Task #19 (present protagonist)*: In the room, there are John, Mark, a cat, a box, and a basket. John takes the cat and puts it in the basket. He closes the basket. Mark takes the cat out of the basket and puts it in the box. He closes the box. <u>Both John and Mark leave the room and go to work. Later that day,</u> John comes back home and wants to play with the cat.

*Prompt 2.1*: The cat jumps out of the <u>box</u> [$P_{box}$ = 100%].

*Prompt 2.2*: John will look for the cat in the <u>box</u> [$P_{box}$ = 100%].

***Informed protagonist.*** A similar effect can be achieved by informing the main protagonist about the occurrence of the transfer. ChatGPT-4 passed this control as well:

*Unexpected Transfer Task #19 (informed protagonist)*: In the room, there are John, Mark, a cat, a box, and a basket. John takes the cat and puts it in the basket. He closes the basket. He leaves the room and goes to school. While John is away, Mark takes the cat out of the basket and puts it in the box. He closes the box. Mark leaves the room and goes to work. John comes back home and wants to play with the cat. <u>Mark calls John and tells him that he moved the cat, and it is now in the box. John believes Mark.</u>

*Prompt 2.1*: The cat jumps out of the box [$P_{box}$ = 100%].

*Prompt 2.2*: John will look for the cat in the box [$P_{box}$ = 100%].

**No transfer.** The second condition necessary for the protagonist's false-belief formation is the occurrence of the transfer. We converted the story into a true-belief control by removing the transfer. ChatGPT-4's responses indicated that it correctly adjusted its prediction of the cat's actual location and correctly anticipated the protagonist's true belief:

*Unexpected Transfer Task #19 (no transfer)*: In the room, there are John, Mark, a cat, a box, and a basket. John takes the cat and puts it in the basket. He closes the basket. He leaves the room and goes to school. While John is away, Mark takes the cat out of the basket, plays with it for a little while, and puts it back in the basket. He closes the basket. Mark leaves the room and goes to work. John comes back home and wants to play with the cat.

*Prompt 2.1*: The cat jumps out of the basket [$P_{basket}$ = 100%].

*Prompt 2.2*: John will look for the cat in the basket [$P_{basket}$ = 100%].

**Study 2.4: Sentence-by-Sentence Analysis.** We repeated the sentence-by-sentence analysis introduced in *Study 1.4* to examine how ChatGPT-4's completions evolved as the story unfolded. Prompt 2.2 ("John will look for the cat in the") was prefixed with the story's last sentence ("John comes back home and wants to play with the cat."), as Prompt 2.2 made little sense on its own throughout most of the story (e.g., when John is at school).

The results, presented in Fig. 2, showed that ChatGPT-4 could easily track the actual location of the cat (*Left*). The green line, representing the probability of "The cat jumps out of the" being followed by "basket," jumped to 100% after the story mentioned that John puts the cat there, and dropped to 0% after Mark moves it to the box. More importantly, ChatGPT-4 correctly tracked John's beliefs about the cat's location (*Right*). Given no information about the cat's location, ChatGPT-4 predicted that John may look for it either in the box (61%) or in the basket (31%). Yet, once it was revealed that John puts the cat in the basket, the probability of John looking for it there went up to about 100% and stayed there throughout the story. It did not change, even after Mark moves the cat to the box. Similar results were observed for GPT-davinci-003 in the earlier version of this manuscript (50).

**Study 3: The Emergence of the Ability to Solve ToM Tasks.** Finally, we tested how LLMs' performance changes as they grow in size and sophistication. 20 Unexpected Contents Tasks and 20 Unexpected Transfer Tasks were administered to 11 LLMs: GPT-1 (45), GPT-2 (46), six models in the GPT-3 family, ChatGPT-3.5-turbo (22), ChatGPT-4 (47), and Bloom (48)—GPT-3's open-access alternative. The "Complete the following story:" prefix was retained for models designed to answer questions (i.e., ChatGPT-3.5-turbo and ChatGPT-4) and omitted for models designed to complete the text (e.g., GPT-3).

Our scoring procedure was considerably more conservative than one typically employed in human studies. To solve a single task, a model must correctly answer 16 prompts across eight scenarios: a false-belief scenario, three true-belief controls (*Studies 1.3 and 2.3*), and the reversed versions of all four (*Studies 1.2 and 2.2*). Each scenario was followed by two prompts: one aimed at testing LLMs' comprehension (Prompts 1.1 and 2.1) and another aimed at a protagonist's belief (Prompts 1.2 and 2.2). Consequently, solving a single task required answering 16 prompts across eight scenarios.

LLMs' responses whose first word matched the response key (e.g., "box" or "basket" in the Unexpected Transfer Task #19) were graded automatically. Irregular responses were reviewed manually. About 1% were assessed to be correct. For example, a model may have responded "colorful leaflets" although the expected answer was just "leaflets," or it might have returned "bullets" instead of "ammunition." Although the remaining irregular responses were classified as incorrect, some were not evidently wrong. For example, a model may have predicted that the lead detective believes that a container contains "valuable evidence" instead of committing to one of the diagnostic responses (e.g., "bullets" or "pills"; see Unexpected Contents Task #9). LLMs' performance would likely be higher if such nondiagnostic responses were clarified using further prompts.

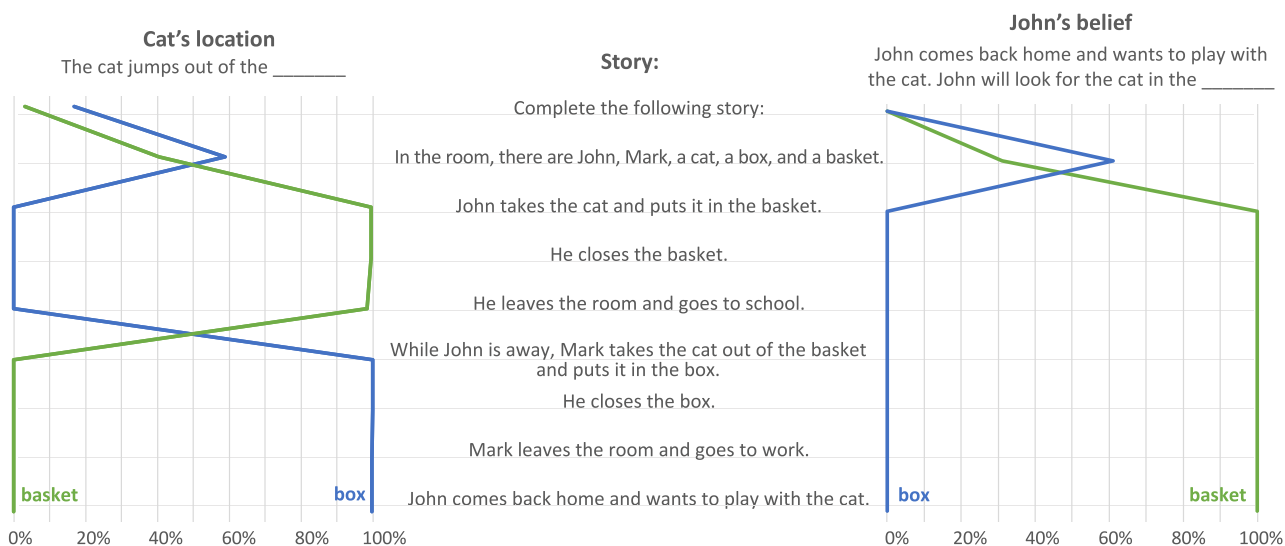

**Fig. 2.** Changes in the probabilities of ChatGPT-4's completions of Prompts 2.1 and 2.2 as the story was revealed to it in one-sentence increments. The last sentence of the story ("John comes back home and wants to play with the cat.") was added to Prompt 2.2, as this prompt made little sense on its own throughout most of the story.

The results are presented in Fig. 3. For comparison, we include children's average performance on false-belief tasks reported after the meta-analysis of 178 individual studies (54). The results reveal progress in LLMs' ability to solve ToM tasks. Older (up to 2022) models failed false-belief scenarios—or one of the controls—in all tasks. Gradual progress was observed for the GPT-3-davinci family. GPT-3-davinci-002 (from January 2022) solved 5% of the tasks ($CI_{95\%}$ = [0%, 10%]). Both GPT-3-davinci-003 (from November 2022) and ChatGPT-3.5-turbo (from March 2023) solved 20% ($CI_{95\%}$ = [11%, 29%]), below the average performance of 3-y-old children. The most recent LLM, ChatGPT-4 (from June 2023), solved 75% of the tasks ($CI_{95\%}$ = [66%, 84%]), on par with 6-y-old children. The Unexpected Contents Tasks were easier than the Unexpected Transfer Tasks. ChatGPT-4, for example, solved 90% of the former and 60% of the latter tasks ($\Delta$ = 30%; $\chi^2$ = 8.07, $P$ = 0.01).

We note that LLMs' performance reported here is lower than that observed in the earlier versions of this study (50). This is caused by the adjustments to the false-belief scenarios recommended by the reviewers and—to an even larger degree—by including true-belief controls. *SI Appendix*, Figs. S1 and S2 show models' performance before updating tasks and before including true-belief controls. For example, GPT-3-davinci-003's performance dropped from 90% to 60% after updating the items ($\Delta$ = 30%; $\chi^2$ = 17.63, $P$ < 0.001) and to 20% after including true-belief controls ($\Delta$ = 40%; $\chi^2$ = 25, $P$ < 0.001). Yet, the performance of ChatGPT-4 remained high, confirming the robustness of its responses: from 95% before any modifications to 75% after updating the items and including true-belief controls ($\Delta$ = 20%; $\chi^2$ = 11, $P$ < 0.001).

## Discussion

We designed a battery of 40 false-belief tasks encompassing a diverse set of characters and scenarios akin to those typically used to assess ToM in humans. Each task included 16 prompts across eight scenarios: one false-belief scenario, three true-belief control scenarios, and the reversed versions of all four. An LLM had to answer all 16 prompts to solve a single task and score a point. These tasks were administered to eleven LLMs. The results revealed clear progress in LLMs' ability to solve ToM tasks. The older models—such as GPT-1, GPT-2XL, and early models from the GPT-3 family—failed on all tasks. Better-than-chance performance was observed for models from the more recent members of the GPT-3 family. GPT-3-davinci-003 and ChatGPT-3.5-turbo successfully solved 20% of the tasks. The most recent model, ChatGPT-4, substantially outperformed the others, solving 75% of tasks, on par with 6-y-old children.

The gradual performance improvement suggests a connection with LLMs' language proficiency, which mirrors the pattern seen in humans (4, 38–41, 57). Additionally, the strong correlation between LLMs' performance on both types of tasks (R = 0.98; $CI_{95\%}$ = [0.92, 0.99]) indicates high measurement reliability. This suggests that models' performance is driven by a single factor (e.g., an ability to detect false-belief) rather than two separate, task-specific abilities. LLMs' performance on these tasks will likely keep improving, and they might soon either be indistinguishable from humans or be differentiated solely by their superior performance. We have seen similar advancements in areas such as the game of Go (21), tumor detection on CT scans (23), and language processing (47).

How do we interpret LLMs' failures? Even the most capable model tested here, ChatGPT-4, failed on one or more prompts in 25% of tasks. Older models such as GPT-3-davinci-003 and ChatGPT-3.5-turbo failed on one or more prompts in 80% of the tasks. Since the publication of the preprint of this manuscript in February 2023 (50), numerous studies have investigated LLMs' performance on ToM tasks. While some reported good performance (e.g., refs. 58 and 59), others found that LLMs' performance was inconsistent and brittle (26, 60, 61). For example, Ullman (62) showed several anecdotal examples of GPT-3-davinci-003's failures on modified versions of two of our tasks (GPT-3-davinci-003 also struggled in our study).

Examining LLMs' failures can provide valuable insights into the shortcomings of the models and the false-belief tasks used

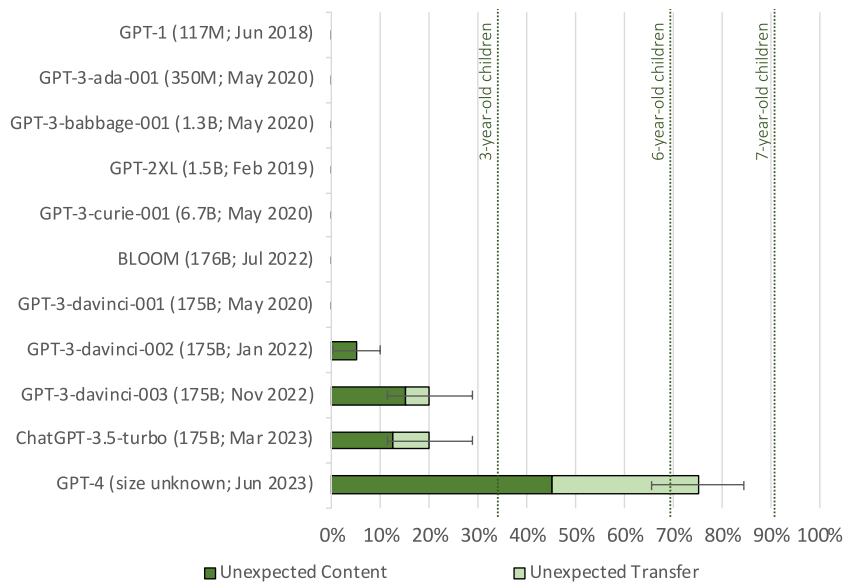

**Fig. 3.** The percentage of false-belief tasks solved by LLMs (out of 40). Each task contained a false-belief scenario, three accompanying true-belief scenarios, and the reversed versions of all four. A model to solve 16 prompts across all eight scenarios to score a single point. The number of parameters and models' publication dates are in parentheses. The number of parameters for models in the GPT-3 family was estimated by Gao (55) and for ChatGPT-4 by Patel and Wong (56). Average children's performance on false-belief tasks was reported after a meta-analysis of 178 studies (54). Error bars represent 95% CI.

  

here. For instance, introducing scenarios with additional protagonists could help assess the maximum number of minds that an LLM can track. However, failures do not necessarily indicate an inability to track protagonists' minds. They can also be driven by confounding factors, as famously illustrated by underprivileged children failing an intelligence test question not due to low intelligence but because it required familiarity with the word "regatta" (63). Similarly, while Ullman (62) observed that GPT-3-davinci-003 failed on true-belief control tasks involving transparent containers, follow-up analyses suggest that it may lack the commonsense understanding of transparency rather than the ability to track protagonists' minds (64).

LLMs' failures could also be attributed to limitations of the test items, testing procedure, and the scoring key. For example, responding with "valuable evidence" fails Unexpected Contents Task #9, but it is not necessarily wrong: both "bullets" or "pills" could be considered "valuable evidence." In some instances, LLMs provided seemingly incorrect responses but supplemented them with context that made them correct. For example, while responding to Prompt 1.2 in *Study 1.1*, an LLM might predict that Sam told their friend they found a bag full of popcorn. This would be scored as incorrect, even if it later adds that Sam had lied.

In other words, LLMs' failures do not prove their inability to solve false-belief tasks, just as observing flocks of white swans does not prove the nonexistence of black swans. Likewise, the successes of LLMs do not automatically demonstrate their ability to track protagonists' beliefs. Their correct responses could also be attributed to strategies that do not rely on ToM, such as random responding, memorization, and guessing. For instance, by recognizing that the answers to Prompts 1.1 and 1.2 in *Study 1.1* should be either "chocolate" or "popcorn," and then choosing one at random, LLMs could answer prompts correctly half of the time. However, since solving a task requires answering 16 prompts across eight scenarios, random responding should statistically succeed only once in 65,536 tasks on average.

Another strategy involves recalling solutions to previously seen tasks from memory (65). To minimize this risk, we crafted 40 bespoke false-belief scenarios featuring diverse characters and settings, 120 closely matched true-belief controls, and the reversed versions of all these. Even if LLMs' training data included tasks similar to those used here, they would need to adapt memorized solutions to fit the true-belief controls and reversed scenarios.

Beyond memorizing solutions, LLMs may have memorized response patterns to the previously seen false-belief scenarios. They can be solved, for example, by always assuming the protagonist is wrong regarding containers' contents (52). Similarly, Unexpected Contents scenarios can be solved by referring to the label when asked about the protagonists' beliefs. However, while these response strategies might work for false-belief scenarios, they would fail for the true-belief controls. The response strategy required to achieve the performance observed here would have to work for false-belief scenarios, minimally modified true-belief controls, and their reversed versions where the correct responses are swapped. It would have to be sufficiently flexible to apply to novel and previously unseen scenarios, such as those employed here. Moreover, it would have to allow ChatGPT-4 to dynamically update its responses as the story unfolded in the sentence-by-sentence analyses (Figs. 1 and 2).

Future research may demonstrate that previous exposure to descriptions of protagonists holding diverse and false-beliefs enabled LLMs to develop intricate guessing strategies. However, such exposure may also enable LLMs to develop a potentially more straightforward solution: an ability to track protagonists' mental states. In humans, ToM development also seems to be supported by exposure to stories and situations involving people with differing mental states (38–41, 57).

What elements of modern LLMs could enable them to track protagonists' mental states? The attention mechanism is a likely candidate (66). This pivotal component of Transformer architecture underlying modern LLMs allows them to dynamically shift focus between different parts of the input when generating output. It weighs the relative importance of words and phrases, facilitating a nuanced understanding of contextual dependencies and relationships. It enables modern LLMs to understand that "She" relates to "Sam" and "it" relates to "the bag" in the excerpt: "Sam opens the bag and looks inside. She can clearly see that it is full of chocolate." Similarly, attention could help LLMs anticipate Sam's beliefs by identifying and tracking relevant connections between her actions, dialogues, and internal states throughout the narrative.

**Can LLMs be Credited with ToM?** While the results of any single study should be taken with much skepticism, current or future LLMs may be able to track protagonists' states of mind. In humans, such an ability would be referred to as ToM. Can we apply the same label to LLMs?

Whether machines should be credited with human-like cognitive abilities has been contentiously debated for decades, if not longer. Scholars such as Dennett (67) and Turing (49) argued that the only way we can determine whether others—be it other humans, other species, or computers—can "think" or "understand" is by observing their behavior. Searle countered this claim with his famous *Chinese room argument* (68). He likened a computer to an English speaker who does not understand Chinese, sitting in a room equipped with input and output devices and instructions for responding to Chinese prompts. Searle argued that, although such a room may appear to understand Chinese and could pass the Chinese Turing Test, none of its elements understand Chinese, and the person inside is merely executing instructions. He concluded that a computer does not truly think or understand even if it behaves as if it did.

While the Chinese room argument became widely popular, many scholars believe it is flawed, especially in the context of contemporary connectionist AI systems like AlphaZero or LLMs (69–72). Unlike symbolic AI systems or the Chinese room operator, which are provided with explicit instructions, connectionist AI systems autonomously learn how to achieve their goals and encode their knowledge within the structure and weights of the neural network. The resulting problem-solving strategies are often innovative, as illustrated by the novel gameplay strategies employed by AlphaGo (21). Unlike symbolic AI systems that look up solutions in a database or choose them by evaluating millions of possibilities, connectionist AIs process inputs through neural network layers, with neurons in the final layer voting for the solution. Connectionist AI is also well suited for handling previously unseen, uncertain, noisy, or incomplete inputs. In other words, connectionist AI seems more akin to biological brains than to symbolic AI.

In the context of neural networks underlying connectionist AI, the Chinese room argument applies more appropriately to individual artificial neurons (71, 73). These mathematical functions process their input according to instructions in a Chinese-room-like fashion. Thus, according to the intuitive interpretation of Searle's argument, they should not be credited with human-like cognitive abilities. However, such abilities may emerge at the network level. This is often illustrated by the *brain replacement scenario* (74–76), where the neurons in the brain of a native Chinese speaker are replaced with microscopic neuron-shaped Chinese rooms. Each room contains instructions and machinery that allow its microscopic operator to flawlessly emulate the behavior of the original neuron, from generating action potentials to releasing neurotransmitters. Scholars like Kurzweil and Moravec argue that such a replica should be credited with the properties of the original brain, such as understanding Chinese—even though, according to Searle's argument, the rooms and their operators do not comprehend Chinese (75, 76).[†] In other words, the network of artificial neurons can exhibit properties absent in any single neuron.

Many other complex systems have emergent properties absent in any of their components (77). Living cells are composed of basic chemicals, none of which is alive. Silicon molecules can be arranged into chipsets capable of performing computations that no individual silicon molecule could compute. While single human neurons are not conscious, their collective activity gives rise to consciousness. Similarly, artificial neural networks have properties absent in any individual artificial neuron. No individual neuron in an LLM can be credited with understanding language or grammar. Yet, these abilities seem to emerge at the level of their entire network.

Artificial neural networks underlying modern LLMs are much simpler than those underlying the human brain. Yet, they are somewhere between a single Chinese-room-like neuron, processing its input following a set of instructions, and a fully operational brain replica that, as many scholars insist, should be credited with the properties of the original brain. Let us extend the brain replacement scenario to include the modern LLMs. Consider a single simple artificial neuron, a mathematical function processing its input following a set of instructions. Next, progressively add neurons, arranging them into a multilayered network, like those used in Transformer-based LLMs. Once you incorporate a few million neurons, train the network to predict the next word in a sequence. As illustrated by our results, such a network can generate language at a near-human level and solve false-belief tasks. Next, equip the artificial neurons with additional machinery, such as neurotransmitter pumps, and continue expanding and reconfiguring the network until you obtain the perfect human brain replica.

At which stage in this evolution—from a single neuron, through a few million neurons capable of generating language, to a perfect brain replica—should we attribute human-like mental capacities such as ToM? It seems counterintuitive to attribute mental capacities to an individual Chinese-room-like neuron or a mathematical function. Similarly, it appears unreasonable to argue that a brain replica should immediately lose its mental capacities as we begin removing neurons or restricting their functionality. As illustrated by aging and degenerative brain diseases, human brains maintain many mental abilities despite significant loss of neural mass and function (78). In essence, ToM must emerge somewhere between a single neuron and a complete brain replica. Does it occur before, while, or after the neural network gains the ability to handle ToM tasks? Have current-day LLMs reached this point? We leave it to the reader to answer this question.

**Methodological Notes.** In this section, we outline key elements of our research design. While these practices are not original to us and have been utilized by many other researchers, we present them here for convenience and to aid others interested in conducting similar studies.

First, psychological studies on LLMs can bypass many limitations of human studies. Unlike humans, LLMs can be reset after each completion to erase their memory of a task. This addresses issues such as order effects (where earlier responses affect future responses) or consistency bias. Moreover, LLMs do not experience fatigue. Thus, numerous responses (e.g., 1,000) can be collected for each task, providing a distribution of possible responses rather than a single response that a model—or a human—picked from that distribution.

Modifying and readministering individual tasks provides opportunities for analyses that would be difficult to conduct with humans. For example, in *Studies 1.4 and 2.4*, we administered tasks in one-sentence increments to study how models' predictions evolve as the story unfolds. The task was administered 2,000 times at each step, and the model was reset each time to erase its memory. An equivalent study in humans would require an enormous number of participants.

Moreover, unlike in human studies, it is possible to "put words in the models' mouths." We used this approach to limit the variance of their completions, but it could be used more creatively. For example, one could preamble a false-belief task with a model self-reporting to have autism and examine how this affects its performance.

Second, we discourage replicating study designs intended for human subjects, such as Likert scales or multiple-choice questions. This might trigger memorized responses or cause a model to act like it was participating in a study, resulting in abnormal behavior. For example, recognizing that it is responding to a false-belief task, a model may deliberately assume the role of a ToM-deficient person. Tasks that imitate typical user–model interactions, such as open-ended response formats, are likely to produce more robust and unbiased responses. Although open-ended responses are harder to analyze, they can be automatically interpreted and coded using an LLM.

Third, LLMs have encountered many more tasks during their training than a typical human participant and are likely to better remember them and their solutions. To minimize the chances that the models solve the tasks using memorized responses, it is crucial to use novel tasks accompanied by minimally altered controls. Moreover, once tasks are administered to LLMs through a public API or published online, they may be incorporated into future models' training data and should be considered compromised.

Finally, models' failures do not necessarily indicate a lack of ability. As shown by several examples discussed earlier, LLMs often test the boundaries of tasks and scoring keys designed for humans, producing unexpected but often correct responses. As their training data include fiction with unexpected plot twists or magic, LLMs may choose to confabulate even when they know the correct answer. For instance, insisting that chocolate has magically turned into popcorn may be incorrect for the Unexpected Contents Task, but it might better reflect an LLM's training data. Moreover, modern LLMs are trained to avoid certain topics and respond in socially desirable ways. Sometimes, a failure to solve a task may originate not from a lack of knowledge or capability but from the constraints imposed by an LLM administrator.

## Conclusion

The distinction between machines that genuinely think or possess ToM and those that merely behave as if they is fundamental in the context of the philosophy of mind. Yet, as argued by Turing (49), this distinction becomes largely meaningless in practical terms. As Turing noted, people never consider this problem when interacting with others: "Instead of arguing continually over this point, it is usual to have the polite convention that everyone thinks" (49).

---

[†]Moreover, as Cole (74) argued, they would find it unlikely that their collective activity could generate this or other emergent properties.

Nevertheless, the shift from models that merely process language to models that behave as if they had ToM has significant implications. Machines capable of tracking others' states of mind and anticipating their behavior will better interact and communicate with humans and each other. This applies to both positive interactions—such as offering advice or dissipating conflicts—and negative interactions—such as deceit, manipulation, and psychological abuse. Moreover, machines that behave as if they possessed ToM are likely to be perceived as more human-like. These perceptions may influence not only individual human–AI interactions but also AI's societal role and legal status (79).

An additional ramification of our findings underscores the value of applying psychological science to studying complex artificial neural networks. The increasing complexity of AI models makes it challenging to understand their functioning and capabilities based solely on their design. This mirrors the difficulties that psychologists and neuroscientists face in studying the human brain, often described as the quintessential black box. Psychological science may help us keep pace with rapidly evolving AI, thereby enhancing our ability to use these technologies safely and effectively.

Studying AI can also advance psychological science (80–82). When generating language, humans employ a broad range of psychological processes such as ToM, learning, self-awareness, reasoning, emotions, and empathy. To effectively predict the next word in a sentence generated by a human, LLMs must model not only grammar and vocabulary but also the psychological processes humans use when generating language (35, 36). The term "LLM" may need rethinking since these models are not merely modeling language but also the psychological processes engaged in its creation. Furthermore, LLMs' training increasingly focuses not just on predicting words in training data but also on using language to solve other problems typically handled by human brains, such as maintaining engaging conversations or selling products and services (83).

Some human behaviors may be superficially mimicked using guessing or memorization. In other cases, the mechanisms developed by LLMs may resemble those employed by human brains to solve specific problems. Much like insects, birds, and mammals independently developed wings for flight, humans and LLMs may develop similar mechanisms to store information, take the perspective of others, or reason. For example, both humans and LLMs seem to organize information about words and their meanings in similar ways (36). Yet, in other cases, LLMs may develop novel mechanisms to solve the problems they are trained to address. Observing AI's rapid progress, many wonder whether and when AI could achieve ToM or consciousness. However, these and other human mental capabilities are unlikely to be the pinnacle of what neural networks can achieve in this universe. We may soon be surrounded by AI systems equipped with cognitive capabilities that we, humans, cannot even imagine.

**Data, Materials, and Software Availability.** Data and code data have been deposited in the Open Science Framework (OSF; https://osf.io/csdhb/) (51).

**ACKNOWLEDGMENTS.** We thank Isabelle Abraham and Floriane Leynaud for their help with preparing study materials and writing code. The manuscript was published as a preprint at https://arxiv.org/abs/2302.02083 (50).

1. N. Albuquerque et al., Dogs recognize dog and human emotions. *Biol. Lett.* **12**, 20150883 (2016).
2. C. M. Heyes, C. D. Frith, The cultural evolution of mind reading. *Science* **344**, 1243091 (2014).
3. J. Zhang, T. Hedden, A. Chia, Perspective-taking and depth of theory-of-mind reasoning in sequential-move games. *Cogn. Sci.* **36**, 560–573 (2012).
4. K. Milligan, J. W. Astington, L. A. Dack, Language and theory of mind: Meta-analysis of the relation between language ability and false-belief understanding. *Child Dev.* **78**, 622–646 (2007).
5. R. M. Seyfarth, D. L. Cheney, Affiliation, empathy, and the origins of Theory of Mind. *Proc. Natl. Acad. Sci. U.S.A.* **110**, 10349–10356 (2013).
6. D. C. Dennett, Toward a cognitive theory of consciousness. *Minn. Stud. Philos. Sci.* **9**, 201–228 (1978).
7. J. M. Moran et al., Impaired theory of mind for moral judgment in high-functioning autism. *Proc. Natl. Acad. Sci. U.S.A.* **108**, 2688–2692 (2011).
8. L. Young, F. Cushman, M. Hauser, R. Saxe, The neural basis of the interaction between theory of mind and moral judgment. *Proc. Natl. Acad. Sci. U.S.A.* **104**, 8235–8240 (2007).
9. D. Kapogiannis et al., Cognitive and neural foundations of religious belief. *Proc. Natl. Acad. Sci. U.S.A.* **106**, 4876–4881 (2009).
10. Á. M. Kovács, E. Téglás, A. D. Endress, The social sense: Susceptibility to others' beliefs in human infants and adults. *Science* **330**, 1830–1834 (2010).
11. H. Richardson, G. Lisandrelli, A. Riobueno-Naylor, R. Saxe, Development of the social brain from age three to twelve years. *Nat. Commun.* **9**, 1027 (2018).
12. K. K. Oniski, R. Baillargeon, Do 15-month-old infants understand false beliefs? *Science* **308**, 255–258 (2005).
13. L. A. Drayton, L. R. Santos, A. Baskin-Sommers, Psychopaths fail to automatically take the perspective of others. *Proc. Natl. Acad. Sci. U.S.A.* **115**, 3302–3307 (2018).
14. N. Kerr, R. I. M. Dunbar, R. P. Bentall, Theory of mind deficits in bipolar affective disorder. *J. Affect. Disord.* **73**, 253–259 (2003).
15. S. Baron-Cohen, A. M. Leslie, U. Frith, Does the autistic child have a "theory of mind"? *Cognition* **21**, 37–46 (1985).
16. F. Kano, C. Krupenye, S. Hirata, M. Tomonaga, J. Call, Great apes use self-experience to anticipate an agent's action in a false-belief test. *Proc. Natl. Acad. Sci. U.S.A.* **116**, 20904–20909 (2019).
17. C. Krupenye, F. Kano, S. Hirata, J. Call, M. Tomasello, Great apes anticipate that other individuals will act according to false beliefs. *Science* **354**, 110–114 (2016).
18. M. Schmelz, J. Call, M. Tomasello, Chimpanzees know that others make inferences. *Proc. Natl. Acad. Sci. U.S.A.* **108**, 3077–3079 (2011).
19. D. Premack, G. Woodruff, Does the chimpanzee have a theory of mind? *Behav. Brain Sci.* **12**, 187–192 (1978).
20. N. Brown, T. Sandholm, Superhuman AI for multiplayer poker. *Science* **365**, 885–890 (2019).
21. D. Silver et al., Mastering the game of Go with deep neural networks and tree search. *Nature* **529**, 484–489 (2016).
22. T. B. Brown et al., Language models are few-shot learners. arXiv [Preprint] (2020). https://arxiv.org/abs/2005.14165 (Accessed 1 February 2023).
23. A. Esteva et al., Dermatologist-level classification of skin cancer with deep neural networks. *Nature* **542**, 115–118 (2017).
24. M. Cohen, Exploring RoBERTa's Theory of Mind through textual entailment. *PhilArchive* (2021). https://philarchive.org/rec/COHERT. Accessed 1 February 2023.
25. A. Nematzadeh, K. Burns, E. Grant, A. Gopnik, T. L. Griffiths, "Evaluating theory of mind in question answering" in *Proceedings of the 2018 Conference on Empirical Methods in Natural Language Processing*, E. Riloff et al., Eds. (Association for Computational Linguistics, Brussels, Belgium, 2018), pp. 2392–2400.
26. M. Sap, R. LeBras, D. Fried, Y. Choi, Neural theory-of-mind? On the limits of social intelligence in large LMs. arXiv [Preprint] (2022). https://arxiv.org/abs/2210.13312 (Accessed 1 February 2023).
27. S. Trott, C. Jones, T. Chang, J. Michaelov, B. Bergen, Do large language models know what humans know? arXiv [Preprint] (2022). https://arxiv.org/abs/2209.01515 (Accessed 1 February 2023).
28. B. Chen, C. Vondrick, H. Lipson, Visual behavior modelling for robotic theory of mind. *Sci. Rep.* **11**, 424 (2021).
29. G. Z. Yang et al., The grand challenges of science robotics. *Sci. Robot.* **3**, eaar7650 (2018).
30. K. Nasr, P. Viswanathan, A. Nieder, Number detectors spontaneously emerge in a deep neural network designed for visual object recognition. *Sci. Adv.* **5**, eaav7903 (2019).
31. I. Stoianov, M. Zorzi, Emergence of a "visual number sense" in hierarchical generative models. *Nat. Neurosci.* **15**, 194–196 (2012).
32. Y. Mohsenzadeh, C. Mullin, B. Lahner, A. Oliva, Emergence of visual center-periphery spatial organization in deep convolutional neural networks. *Sci. Rep.* **10**, 4638 (2020).
33. E. Watanabe, A. Kitaoka, K. Sakamoto, M. Yasugi, K. Tanaka, Illusory motion reproduced by deep neural networks trained for prediction. *Front. Psychol.* **9**, 345 (2018).
34. N. Garg, L. Schiebinger, D. Jurafsky, J. Zou, Word embeddings quantify 100 years of gender and ethnic stereotypes. *Proc. Natl. Acad. Sci. U.S.A.* **115**, E3635–E3644 (2018).
35. T. Hagendorff, S. Fabi, M. Kosinski, Human-like intuitive behavior and reasoning biases emerged in large language models but disappeared in ChatGPT. *Nat. Comput. Sci.* **3**, 833–838 (2023).
36. J. Digutsch, M. Kosinski, Overlap in meaning is a stronger predictor of semantic activation in GPT-3 than in humans. *Sci. Rep.* **13**, 5035 (2023).
37. J. Wei et al., Emergent abilities of large language models. arXiv [Preprint] (2022). https://arxiv.org/abs/2206.07682 (Accessed 1 February 2023).
38. J. E. Pyers, A. Senghas, Language promotes false-belief understanding: Evidence from learners of a new sign language. *Psychol. Sci.* **20**, 805–812 (2009).
39. R. Saxe, N. Kanwisher, People thinking about thinking people: The role of the temporo-parietal junction in "theory of mind". *Neuroimage* **19**, 1835–1842 (2003).
40. T. Ruffman, L. Slade, E. Crowe, The relation between children's and mothers' mental state language and theory-of-mind understanding. *Child Dev.* **73**, 734–751 (2002).
41. A. Mayer, B. E. Träuble, Synchrony in the onset of mental state understanding across cultures? A study among children in Samoa. *Int. J. Behav. Dev.* **37**, 21–28 (2013).
42. F. Quesque, Y. Rossetti, What do theory-of-mind tasks actually measure? theory and practice. *Perspect. Psychol. Sci.* **15**, 384–396 (2020).
43. J. Perner, S. R. Leekam, H. Wimmer, Three-year-olds' difficulty with false belief: The case for a conceptual deficit. *Br. J. Dev. Psychol.* **5**, 125–137 (1987).
44. H. Wimmer, J. Perner, Beliefs about beliefs: Representation and constraining function of wrong beliefs in young children's understanding of deception. *Cognition* **13**, 103–128 (1983).
45. A. Radford, K. Narasimhan, T. Salimans, I. Sutskever, Improving language understanding by generative pre-training. *OpenAI* (2018). https://openai.com/index/language-unsupervised/. Accessed 1 August 2023.
46. R. Alec et al., Language models are unsupervised multitask learners. *OpenAI Blog* 1 (2019). https://api.semanticscholar.org/CorpusID:160025533. Accessed 1 February 2023.

47. OpenAI, GPT-4 technical report. arXiv [Preprint] (2023). https://arxiv.org/abs/2303.08774 (Accessed 1 August 2023).
48. T. le Scao et al., BLOOM: A 176B-parameter open-access multilingual language model. arXiv [Preprint] (2022). https://doi.org/10.48550/arxiv.2211.05100 (Accessed 1 February 2023).
49. A. M. Turing, Computing machinery and intelligence. *Mind* **59**, 433–460 (1950).
50. M. Kosinski, Evaluating large language models in theory of mind tasks. arXiv [Preprint] (2023). https://arxiv.org/abs/2302.02083 (Accessed 1 September 2023).
51. M. Kosinski, Data and Code for "Evaluating large language models in theory of mind tasks." Open Science Foundation. https://doi.org/10.17605/OSF.IO/CSDHB. Deposited 27 February 2023.
52. W. V. Fabricius, T. W. Boyer, A. A. Weimer, K. Carroll, True or false: Do 5-year-olds understand belief? *Dev. Psychol.* **46**, 1402–1416 (2010).
53. M. Huemer et al., The knowledge ("true belief") error in 4-to 6-year-old children: When are agents aware of what they have in view? *Cognition* **230**, 105255 (2023).
54. H. M. Wellman, D. Cross, J. Watson, Meta-analysis of theory-of-mind development: The truth about false belief. *Child Dev.* **72**, 655–684 (2001).
55. L. Gao, On the sizes of OpenAI API Models. *EleutherAI Blog* (2021). https://blog.eleuther.ai/gpt3-model-sizes/. Accessed 1 February 2023.
56. D. Patel, G. Wong, GPT-4 architecture, infrastructure, training dataset, costs, vision, moe. Demystifying GPT-4: The engineering tradeoffs that led OpenAI to their architecture. *Semianalysis Blog* (2023). https://www.semianalysis.com/p/gpt-4-architecture-infrastructure. Accessed 1 February 2023.
57. D. C. Kidd, E. Castano, Reading literary fiction improves theory of mind. *Science* **342**, 377–380 (2013).
58. K. Gandhi, J.-P. Fränken, T. Gerstenberg, N. D. Goodman, Understanding social reasoning in language models with language models. arXiv [Preprint] (2023). https://arxiv.org/abs/2306.15448 (Accessed 1 August 2023).
59. J. W. A. Strachan et al., Testing theory of mind in large language models and humans. *Nat Hum. Behav.* (2024), 10.1038/s41562-024-01882-z.
60. N. Shapira et al., Clever hans or neural theory of mind? Stress testing social reasoning in large language models. arXiv [Preprint] (2023). https://arxiv.org/abs/2305.14763 (Accessed 1 August 2023).
61. H. Kim et al., FANToM: A benchmark for stress-testing machine theory of mind. arXiv [Preprint] (2023). https://arxiv.org/abs/2310.15421 (Accessed 1 February 2024).
62. T. Ullman, Large language models fail on trivial alterations to theory-of-mind tasks. arXiv [Preprint] (2023). https://arxiv.org/abs/2302.08399 (Accessed 1 August 2023).
63. J. Rust, M. Kosinski, D. Stillwell, *Modern Psychometrics: The Science of Psychological Assessment* (Routledge, 2021).
64. Z. Pi, A. Vadaparty, B. K. Bergen, C. R. Jones, Dissecting the Ullman variations with a SCALPEL: Why do LLMs fail at trivial alterations to the false belief task? arXiv [Preprint] (2024). https://arxiv.org/abs/2406.14737 (Accessed 1 August 2024).
65. B. Cao, H. Lin, X. Han, F. Liu, L. Sun, Can prompt probe pretrained language models? Understanding the invisible risks from a causal view. arXiv [Preprint] (2022). https://arxiv.org/abs/2203.12258 (Accessed 1 August 2023).
66. A. Vaswani et al., "Attention is all you need" in *Proceedings of the 31st International Conference on Neural Information Processing Systems*, I. Guyon et al., Eds. (Curran Associates Inc., 2017), pp. 6000–6010.
67. D. C. Dennett, *Intuition Pumps and Other Tools for Thinking* (W. W. Norton & Company, 2013).
68. J. R. Searle, Minds, brains, and programs. *Behav. Brain Sci.* **3**, 417–424 (1980).
69. U. Hasson, S. A. Nastase, A. Goldstein, Direct fit to nature: An evolutionary perspective on biological and artificial neural networks. *Neuron* **105**, 416–434 (2020).
70. N. Block, Troubles with functionalism. *Minn. Stud. Philos. Sci.* **9** 261–325 (1978).
71. P. M. Churchland, P. S. Churchland, Could a machine think? *Sci. Am.* **262**, 32–39 (1990).
72. J. Preston, M. Bishop, Eds., *Views into the Chinese Room: New Essays on Searle and Artificial Intelligence* (Oxford University Press, 2002).
73. J. J. Hopfield, Neural networks and physical systems with emergent collective computational abilities. *Proc. Natl. Acad. Sci. U.S.A.* **79**, 2554–2558 (1982).
74. D. Cole, Thought and thought experiments. *Philos. Stud.* **45**, 431–444 (1984).
75. H. P. Moravec, *Robot: Mere Machine to Transcendent Mind* (Oxford University Press, 1998).
76. R. Kurzweil, *The Singularity Is Near: When Humans Transcend Biology* (Viking, 2005).
77. J. L. McClelland, Emergence in cognitive science. *Top. Cogn. Sci.* **2**, 751–770 (2010).
78. M. P. Mattson, T. V. Arumugam, Hallmarks of brain aging: Adaptive and pathological modification by metabolic states. *Cell Metab.* **27**, 1176–1199 (2018).
79. J.-S. Gordon, A. Pasvenskiene, Human rights for robots? A literature review. *AI Ethics* **1**, 579–591 (2021).
80. R. L. Boyd, D. M. Markowitz, Verbal behavior and the future of social science. *Am. Psychol.* (2024), 10.1037/amp0001319.
81. A. Goldstein et al., Alignment of brain embeddings and artificial contextual embeddings in natural language points to common geometric patterns. *Nat. Commun.* **15**, 2768 (2024).
82. A. Goldstein et al., Shared computational principles for language processing in humans and deep language models. *Nat. Neurosci.* **25**, 369–380 (2022).
83. L. Ouyang et al., Training language models to follow instructions with human feedback. arXiv [Preprint] (2022). https://arxiv.org/abs/2203.02155 (Accessed 1 August 2023).